\UseRawInputEncoding
\documentclass[11pt]{article}

\usepackage[final]{acl}

\usepackage{times}
\usepackage{latexsym}

\usepackage[T1]{fontenc}

\usepackage[utf8]{inputenc}

\usepackage{microtype}

\usepackage{inconsolata}

\usepackage{graphicx}

\usepackage{makecell}
\usepackage{booktabs}
\usepackage{multirow}
\usepackage{caption}
\usepackage{subcaption}
\usepackage{amsmath}

\usepackage{listings}
\usepackage{xcolor}

\lstdefinestyle{mystyle}{
    basicstyle=\ttfamily\small,
    keywordstyle=\color{blue},
    commentstyle=\color{gray},
    stringstyle=\color{red!70!black},
    showstringspaces=false,
    breaklines=true,
    frame=single,
    numbers=left,
    numberstyle=\tiny\color{gray},
}
   \usepackage{pgfplots}
   \pgfplotsset{compat=1.18}

\usepackage{adjustbox}

\definecolor{todo}{rgb}{1,0,0}

 \definecolor{colW0}{HTML}{4C72B0}
 \definecolor{colW1}{HTML}{DD8452}
 \definecolor{colW2}{HTML}{55A868}
 \definecolor{colW3}{HTML}{C44E52}
 \definecolor{colFS1}{HTML}{4C72B0}
 \definecolor{colFS2}{HTML}{DD8452}
 \definecolor{colFS3}{HTML}{55A868}
 \definecolor{colFS4}{HTML}{C44E52}
 \pgfplotsset{
   barfig/.style={
     ybar, bar width=18pt, enlarge x limits=0.25,
     ylabel={Mean F1 Score}, ylabel style={font=\normalsize},
     xlabel style={font=\normalsize},
     ymajorgrids=true, grid style={dashed, gray!40},
     tick label style={font=\normalsize}, axis line style={gray!60},
     width=\linewidth, height=7cm,
     nodes near coords, nodes near coords align={vertical},
     every node near coord/.append style={font=\small,
       /pgf/number format/fixed, /pgf/number format/precision=3},
   },
   barfig_std/.style={barfig, ymin=0.30, ymax=0.55},
   barfig_esc/.style={barfig, ymin=0.45, ymax=0.80},
 }

\usepackage[breakable,listings]{tcolorbox}
\tcbuselibrary{skins}
\newtcblisting{promptbox}{
  enhanced,
  colback=gray!5,
  colframe=black!60,
  listing only,
  title=System Prompt,
  boxsep=2pt,
  left=4pt,
  right=4pt,
  top=2pt,
  bottom=2pt,
  listing options={
    basicstyle=\ttfamily\fontsize{6pt}{7pt}\selectfont,
    breaklines=true,
    breakatwhitespace=false,
    breakautoindent=true,
    columns=fullflexible,
    keepspaces=true,
    breakindent=0pt,
    showspaces=false,
    showstringspaces=false,
    postbreak=\mbox{\textcolor{red}{$\hookrightarrow$}\space},
    xleftmargin=0pt,
    xrightmargin=0pt,
  }
}

\title{CUNY at CLPsych 2026: A Pipeline Approach to Classification and Summarization of Mental Health Changes}
    
\author{
    Amirmohammad Ziaei Bideh\textsuperscript{$\dagger$},
    Shameed Charlomar Job\thanks{Equal contribution.}\textsuperscript{$\ddagger$}, \\
    \textbf{Ava Yahyapour\footnotemark[\value{footnote}]}\textsuperscript{$\dagger$},
    \textbf{Alla Rozovskaya}\textsuperscript{$\dagger \ddagger$}
    \\
    \textsuperscript{$\dagger$}Computer Science Department, CUNY Graduate Center \\
    \textsuperscript{$\ddagger$}Linguistics Department, CUNY Graduate Center \\
    \texttt{aziaeibideh@gradcenter.cuny.edu} 
    }
\begin{document}
\maketitle
\begin{abstract}
We describe our submission to the CLPsych~2026 Shared Task on capturing and characterizing mental health changes through social media timeline dynamics. To infer the dominant self-states in posts (Tasks 1.1 and 1.2), we ensemble in-context learning of three open-weight large language models using majority voting. For predicting moments of change in a timeline (Task~2), we train supervised classifiers on features derived from Task~1.1 predictions. To summarize the patterns of mood dynamics and their progression over time within a timeline (Task 3.1), we augment in-context example labels predicted by upstream systems (Tasks 1.1, 1.2, and 2), yielding performance gains over zero-shot and unaugmented in-context learning baselines. Our submission ranked first on Task~1.1, fourth on Task~1.2, fourth on Task~2, and third on Task~3.1.\footnote{The source code for the experiments is available at \url{https://github.com/amirzia/clpsych26-cuny}.}
\end{abstract}

\section{Introduction}
\label{sec:introduction}
Mental health conditions affect a significant portion of the global population~\cite{who-2013-mental-health, nimh-mental-illness}, creating a need for scalable tools to monitor individuals' psychological states over time. Social media platforms offer longitudinal data for tracking how mental states evolve in response to life events and social interactions~\cite{tsakalidis-etal-2022-overview}, and large language models (LLMs) have shown strong potential in supporting such analysis~\cite{yang-etal-2024-mentalllama, chan-etal-2025-prompt}. Recent practice-oriented research further demonstrates how multimodal analysis and AI  can uncover the intrapersonal and interpersonal dynamics underlying therapeutic change~\cite{atzil-slonim-2026-therapeutic-change}. The CLPsych 2026 shared task~\cite{ali026overview} addresses this need by asking participants to track and characterize how users' mental states evolve across longitudinal Reddit timelines.

This paper describes our submission to the CLPsych 2026 shared task~\cite{ali026overview}. Our best submission for Task~1.1 uses an ensemble of seven LLM predictions across three model backbones with subelement-level In-Context Learning (ICL), where annotated training examples are included directly in the prompt to guide predictions. For Task~1.2, our best submission ensembles five predictions using ICL and Retrieval-Augmented Generation (RAG), which retrieves the most semantically similar training posts as in-context demonstrations. For Task~2, we train supervised classifiers on self-state features derived from upstream task predictions, using a Support Vector Machine (SVM) for Switch detection and a Random Forest (RF) for Escalation detection. For Task~3.1, our best submission uses label-augmented ICL, enriching prompts with predicted ABCD subelement and Moment of Change (MoC) labels from upstream tasks. For Task~3.2, we apply a batch-and-merge pipeline to identify recurrent dynamic signatures of improvement and deterioration across timelines. Our submission ranked \textbf{first} on Task~1.1, \textbf{fourth} on Task~1.2, \textbf{fourth} on Task~2, and \textbf{third} on Task~3.1.

The contributions of our work are as follows:

\begin{enumerate}
    \item We show that LLM ensembling via majority voting substantially outperforms single-model baselines for subelement classification (Task~1.1), and that expanding the ensemble beyond five members further improves subelement classification but slightly hurts presence scoring (Task~1.2), suggesting diminishing returns for larger ensembles on ordinal prediction tasks.
    \item  We show that escalation is easier to predict than switch, and traditional supervised classifiers trained on LLM-derived presence scores are competitive for MoC detection (Task~2).
    \item We find that propagating predicted self-state labels from upstream tasks into downstream prompts with larger models yields consistent gains in summary quality over zero-shot prompting with smaller models and standard ICL baselines (Task~3.1).
\end{enumerate}

\section{Shared Task Description}
\label{sec:shared-task}
The shared task is grounded in the  Multimodal Intrapersonal and Interpersonal
Dynamics (MIND) framework \cite{atzil_slonim_2025_mind}. The framework includes widely used psychotherapy constructs on the patient mental state referred to as {self-state}. A self-state is a dominant mode of experience and consists of a combination of Affect, Behavior, Cognition, and Desire (ABCD) components, their adaptivity level and their more fine-grained subcategorization into subelements~\cite{atzil_slonim_2025_mind}.

Task~1.1 aims to identify which predefined ABCD subelements are expressed in a post and how they combine into \textit{adaptive} and \textit{maladaptive} self-states, referred to as the two \textit{valences} of a self-state. 
Task 1.2 requires quantifying  the degree to which each identified self-state is present in the post on a 1-5 scale (referred to as the presence score).  Task 2 involves detecting clinically meaningful MoC within a user timeline,\footnote{A \emph{timeline} is a chronologically ordered collection of posts authored by a single user.} identifying Switches (sudden change in well-being) and Escalations (gradual intensification of mood) \cite{tsakalidis-etal-2022-overview}. Task 3.1 involves generating a structured  summary describing the progression of self-state dynamics within a sequence of posts surrounding an identified change event. Task~3.2 aims to identify recurrent dynamic signatures of improvement and deterioration that recur across multiple sequences and individuals.

\paragraph{Dataset.} The training set contains Reddit timelines from 30 users, annotated with self-states according to the MIND framework. 
Figure~\ref{fig:example-timeline} shows an anonymized excerpt of a sample timeline. The training set also provides gold summaries for sequences, describing the patterns of self-state dynamics. A sequence is a chronologically ordered list of posts within a timeline that culminates in an MoC. We randomly hold out 10 training timelines as our validation set for selecting the best models for submission. The 20 timelines are used for training and providing in-context examples.

\section{Related Work}
Previous editions of the CLPsych shared task have attracted a wide range of system submissions. In CLPsych 2022 \cite{tsakalidis-etal-2022-overview}, team \texttt{BLUE} \cite{bucur2022capturing} experimented with several text representation methods, and their best system consisted of an ensemble of machine learning (ML) classifiers. Team \texttt{WResearch} \cite{bayram-benhiba-2022-emotionally} adopted a pipeline approach in which pre-trained BERT \cite{devlin2019bert} was used to compute emotion and sentiment scores that were passed as input features to downstream ML models. Team \texttt{UoS} \cite{azim-etal-2022-detecting} achieved competitive results using a bidirectional long short-term memory (Bi-LSTM) network for mood change prediction and suicide risk level assessment.

The following year's edition, CLPsych 2025 \cite{tseriotou-etal-2025-overview}, introduced more challenging subtasks such as evidence span detection and summarization. Team \texttt{uOttawa} \cite{chan-etal-2025-prompt} explored various prompt engineering strategies on top of a 70B-parameter LLM and obtained the best score on self-state identification. Team \texttt{BULUSI} \cite{ravenda-etal-2025-evidence} achieved strong results by combining an ensemble with an optimization step on top of LLM predictions. Finally, team \texttt{BLUE} \cite{sandu-etal-2025-capturing} obtained competitive performance on summarization through zero-shot prompting of open-weight LLMs.

\section{Methodology}


This section presents our approaches, which integrate findings from the CLPsych 2025 shared task, as we enhance and refine our strategies.

\subsection{Tasks 1.1 and 1.2}

We adopt a joint prediction setup in which an LLM is tasked with predicting both the subelements and their presence scores in a single pass. The default system prompt (Figure~\ref{fig:sys-prompt-1.1}) contains a detailed description of the MIND framework, the definitions of self-states, the characteristics of each subelement, and the criteria for assigning presence scores. We compare several prompting techniques.

\textbf{Zero-shot prompting.} We use the default system prompt and pass the content of the post to be labeled in the user message.

\textbf{Post-level In-Context Learning (ICL).} Here, $k$ full training posts selected uniformly at random, and their gold subelement annotations are included in the prompt. This exposes the LLM to the surrounding context of each subelement and to the relationships between subelements within a post. A drawback is that coverage of all subelements is not guaranteed, and rare subelements may go unrepresented among the in-context examples. 

\textbf{Post-level ICL with RAG.} Identical to post-level ICL, but the $k$ in-context posts are retrieved by cosine similarity to the test post. Posts are encoded as the L2-normalized \texttt{CLS} embedding from \texttt{BAAI/bge-large-en-v1.5} \cite{bge_embedding}, truncated to 512 tokens.

\textbf{Subelement-level ICL.} In the subelement-level variant of ICL, we append $k$ examples to the definition of each subelement in the system prompt. Each example is a relevant span from a training post that serves as evidence for the corresponding subelement. Note that, in this setting, we do not include the full post. Moreover, the $k$ examples for a given subelement may come from different training posts, offering a more diverse view of how the subelement is expressed.

\textbf{Ensemble.} To reduce noise from any single prediction, we aggregate the outputs of multiple independent LLM runs via majority voting. 

\subsection{Task 2}
 We train separate supervised classifiers for switch and escalation changes using feature sets composed of the subelements and presence scores predicted in Tasks 1.1 and 1.2. Each classifier receives a fixed-size window of posts centered on a target post. The window includes both preceding and following posts; we denote the window that includes the following posts as having foresight. Each post in a sequence is labeled with predictions from the best submissions for Tasks 1.1 and 1.2. We experiment with the following features: the predicted presence for each valence, the absolute difference between presence per valence of the target and subsequent post, count of subelements per valence, and post index. We compare two machine learning algorithms: Support Vector Machine (SVM), and Random Forest. Classifiers are trained on 20 posts during validation (all 30 posts are used during test) and are tuned via grid search
(Appendix~\ref{sec:hyperparams}).  

\subsection{Task 3.1}
We adopt an LLM prompting approach for Task 3.1. The system prompt that we use for the task (Figure~\ref{fig:sys-prompt-3.1}) contains a detailed description of the MIND framework, the definitions of switch and escalation, and the required summarization aspects. We also experimented with a shortened version of this prompt (Figure~\ref{fig:sys-prompt-3.1-short}) but observed negligible difference in performance, so all reported results use the longer prompt. The user prompt contains the chronologically ordered post contents of the test sequence. We compare the following approaches:

\textbf{Zero-shot.} The LLM produces the summary directly from the task description and the post contents, with no in-context examples.

\textbf{ICL.} The system prompt is augmented with $k$ in-context examples. Each example consists of the post contents of a training sequence followed by the corresponding gold summary.

\begin{figure*}[t]
    \centering
    \includegraphics[width=0.63\linewidth]{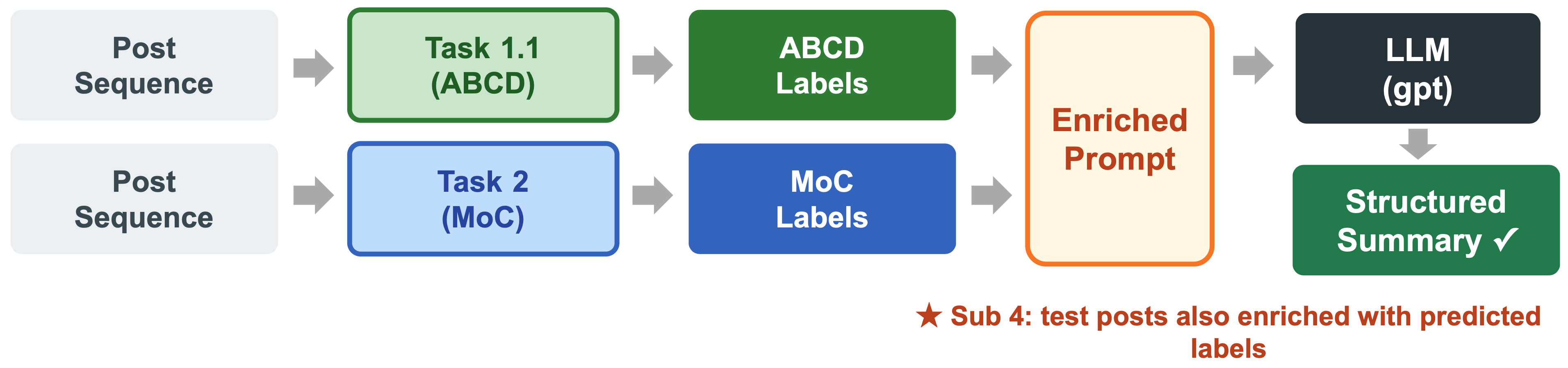}
    \caption{Label-augmented ICL pipeline for Task~3.1. Post sequences are processed through Task~1.1 to produce subelements and through Task~2 to produce MoC labels. Both label sets enrich the prompt fed to the LLM. Submission~3 enriches only the in-context examples with gold labels. Submission~4 additionally enriches the test posts with predicted labels from Tasks~1.1 and~2 at inference time.}
    \label{fig:pipeline}
\end{figure*}

\textbf{Label-augmented ICL.} A pipeline-style extension of ICL in which subelement and change labels (switch or escalation) are included alongside the post contents. We consider two variants: (i) augmenting only the in-context examples with their gold labels, and (ii) additionally augmenting the test posts with predicted labels from our Task~1.1 and Task~2 systems. Figure~\ref{fig:pipeline} illustrates the full pipeline.

\textbf{Summary of summaries.} In this method, a single LLM first summarizes each post individually and then summarizes the sequence from those per-post summaries.
\begin{table*}[t]
\centering
\begin{tabular}{l c c c c c c c c c}
\toprule
\multirow{2}{*}{Method} & \multirow{2}{*}{$k$} & \multicolumn{2}{c}{\texttt{qwen}} & \multicolumn{2}{c}{\texttt{gemma}} & \multicolumn{2}{c}{\texttt{gpt}} & \multicolumn{2}{c}{Ensemble} \\
\cmidrule(lr){3-4} \cmidrule(lr){5-6} \cmidrule(lr){7-8} \cmidrule(lr){9-10}
 &  & F1$\uparrow$ & RMSE$\downarrow$ & F1$\uparrow$ & RMSE$\downarrow$ & F1$\uparrow$ & RMSE$\downarrow$ & F1$\uparrow$ & RMSE$\downarrow$ \\
\midrule
Zero-shot & -- & 0.309 & 1.124 & 0.324 & 0.926 & 0.340 & 0.908 & 0.330 & 0.913 \\
\addlinespace[2pt]
Subelement ICL & 1 & 0.328 & 1.027 & 0.335 & 0.860 & \textbf{0.343} & 0.948 & 0.347 & 0.852 \\
 & 2 & \textbf{0.362} & 1.057 & 0.341 & 0.860 & 0.326 & 0.948 & 0.361 & 0.861 \\
 & 3 & 0.333 & \textbf{0.998} & 0.354 & \textbf{0.831} & 0.339 & \textbf{0.888} & \textbf{0.366} & \textbf{0.787} \\
\addlinespace[2pt]
Post ICL & 1 & 0.331 & 1.072 & 0.330 & 0.877 & 0.311 & 0.939 & 0.335 & 0.867 \\
 & 2 & 0.334 & 1.103 & 0.348 & 0.867 & 0.297 & 0.963 & 0.337 & 0.880 \\
 & 3 & 0.330 & 1.047 & 0.336 & 0.860 & 0.307 & 0.971 & 0.332 & 0.859 \\
\addlinespace[2pt]
RAG & 1 & 0.337 & 1.048 & \textbf{0.368} & 0.920 & 0.309 & 0.914 & 0.330 & 0.898 \\
 & 2 & 0.325 & 1.070 & 0.326 & 0.858 & 0.313 & 0.955 & 0.333 & 0.903 \\
 & 3 & 0.332 & 1.028 & 0.327 & 0.853 & 0.319 & 0.953 & 0.342 & 0.876 \\
\bottomrule
\end{tabular}
\caption{Validation results on Tasks~1.1 and~1.2, averaged over 5 runs. $k$ is the number of in-context examples.}
\label{tab:task1-validation}
\vspace{-0.3cm}
\end{table*}

\subsection{Task 3.2}
Before prompting the LLM to identify recurrent signatures of improvement or deterioration, we filter the 74 gold training summaries via exact string matching, yielding 56 deterioration and 51 improvement sequences. These are passed to the LLM in batches of 10 to produce partial signatures, which are then merged in a final step that identifies patterns common across batches.

\section{Results}
This section provides the experimental results. We use the following open-weight LLMs for prompting: 
google/gemma-3-27b-it (\texttt{gemma}) \cite{gemmateam2025gemma3technicalreport}; Qwen/Qwen3.5-27B (\texttt{qwen}) \cite{qwen3.5}; and openai/gpt-oss-120b (\texttt{gpt}) \cite{openai2025gptoss120bgptoss20bmodel}. 
See Appendix~\ref{sec:appendix-setup} for the setup.

\subsection{Tasks 1.1 and 1.2}

Validation results across these dimensions are summarized in Table~\ref{tab:task1-validation}. On Task~1.1, ICL improves over zero-shot for \texttt{qwen} and \texttt{gemma} but not for \texttt{gpt}. For the ensemble, increasing $k$ generally yields gains, unlike for individual models, and subelement-level ICL outperforms the other approaches. On Task~1.2, \texttt{qwen} attains substantially higher RMSE on presence scores than the other models. The LLM exhibits a clear tendency to overpredict subelements, resulting in a high false positive rate. The validation set contains 292 present subelements (adaptive and maladaptive combined), whereas one of the \texttt{gemma} predictions includes 555.

The test results for our three submissions are reported in Table~\ref{tab:task1-results}. \textbf{Submission~1} uses subelement-level ICL with $k=3$ and \texttt{gemma} as the backbone. \textbf{Submission~2} ensembles 5 predictions from \texttt{qwen} and \texttt{gemma}. \textbf{Submission~3} ensembles 7 predictions drawn from \texttt{qwen}, \texttt{gemma}, and \texttt{gpt}. Refer to Appendix \ref{appendix:sec-ensemble} for more details about the ensemble members.

\paragraph{Findings.} Ensembling improves Task~1.1 macro F1 from 0.416 to 0.431 ($+3.6\%$ relative) and reduces Task~1.2 RMSE from 1.047 to 0.997 ($-4.78\%$ relative) between \textbf{Submissions~1} and \textbf{2}, consistent with the view that aggregating across models and runs reduces variance from sampling noise and from the choice of in-context examples \cite{yang2023one}. \textbf{Submission~3}, which adds two more members and a third backbone, achieves the \textbf{best} Task~1.1 F1 among all CLPsych~2026 participants. For Task~1.2, however, expanding the ensemble from 5 to 7 members slightly hurts performance, likely because the output space is restricted to integers in $\{1, \dots, 5\}$, leaving little headroom for additional members to contribute useful signal.

\begin{table}[th]
  \centering
  \setlength{\tabcolsep}{4pt}
  \begin{tabular}{lcc}
    \toprule
    & \makecell{Task 1.1\\Macro F1 $\uparrow$} & \makecell{Task 1.2\\RMSE $\downarrow$} \\
    \midrule
    Submission 1 & 0.416 & 1.047 \\
    Submission 2 & \underline{0.431} & \textbf{0.997} \\
    Submission 3 & \textbf{0.442} & \underline{1.007} \\
    \midrule
    Official baseline & 0.247 & 1.424 \\
    \bottomrule
  \end{tabular}
  \caption{Test results on Tasks~1.1 (subelement-level macro F1) and 1.2 (RMSE). Submission~3 achieves the best F1 among CLPsych~2026 participants, and Submission~2 achieves the 4th-best RMSE. The official baseline is a one-shot approach with \texttt{LLaMA-3.1-8B-Instruct}.}
  \label{tab:task1-results}
    \vspace{-0.3cm}

\end{table}

 \subsection{Task 2}

\textbf{Submission 1}\footnote{The training data was not fully used in this submission. We provide the correct evaluation on the validation set.} uses Random Forest classifiers with predicted Task~1.1 presence scores as the feature, with window size~1 for switch and~2 for escalation.

\textbf{Submission 2} uses SVM classifiers and expands the feature set with the absolute difference in presence between consecutive posts and the post index, and increases window sizes: switch uses a window size~2 and escalation~3.

\textbf{Submission 3} uses a window size~3 and a larger feature set than the previous submission. For both the switch and escalation model, we create a set of six features per post: presence, absolute presence difference, and subelement count for each valence. While the switch model has a window size~3, we remove foresight so that it does not include posts succeeding our target. This switch model also uses post index as an additional feature. 
Our validation results are reported in Table~\ref{tab:2-val-results}. The test results for our three submissions are reported in Table~\ref{tab:task2-results}. 

\paragraph{Findings.} Our best system (Submission~3) achieved a combined F$_1$ of 0.572, placing us 4th in Task~2 ranking. Across submissions, escalation was consistently easier to predict than switch.  Adding foresight for escalation and refining the window configuration in Submission~3 provided a further improvement, largely driven by a large gain in escalation F$_1$ (0.585 $\to$ 0.714 post-level).

\begin{table}[t]
  \centering
  \setlength{\tabcolsep}{4pt}
  \begin{tabular}{lc}
    \toprule
    & \makecell{Average Macro F1 $\uparrow$} \\
    \midrule
     Submission 1 & 0.279 \\
    Submission 2 & \underline{0.472} \\
    Submission 3 & \textbf{0.572} \\
    \midrule
    Official baseline 1 & 0.272 \\
    Official baseline 2 & 0.365 \\
    Official baseline 3 & \textbf{0.572} \\
    \bottomrule
  \end{tabular}
    \caption{Test results on Task 2. Baseline 1 adopts a zero-shot approach using \texttt{LLaMA-3.1-8B-Instruct}. The second baseline follows a pipeline approach where the predicted labels from Task 1.1 baseline are used to compute the well-being score deterministically to predict moments of change. Baseline 3 finetunes TempoFormer \cite{tseriotou2024tempoformer}.}
  \label{tab:task2-results}
  \vspace{-0.3cm}
\end{table}

\subsection{Task 3.1}

Validation results  are reported in Table~\ref{tab:task3.1-results-validation}. ICL performance generally improves with $k$ across all three backbones. Test results are reported in Table~\ref{tab:task3.1-results-test}. \textbf{Submission~1} uses plain ICL with $k=2$ and \texttt{gpt} as the backbone. \textbf{Submission~2} uses the summary-of-summaries approach with the same backbone, which degrades performance across all metrics: collapsing each post into an intermediate summary appears to discard the fine-grained signal needed to characterize cross-post dynamics. \textbf{Submission~3} is the label-augmented ICL variant, in which the in-context examples carry their gold subelements and change labels. \textbf{Submission~4} additionally augments the test posts with predicted labels from Tasks~1.1 and 2. 

\paragraph{Findings.} Submission 4 yields the best result on three of the four metrics among our submissions confirming that the pipeline signal is useful at inference time as well. Moreover, Submissions 1, 3, and 4 outperform both official baselines across all four metrics, demonstrating that larger models with an ICL approach consistently outperform smaller models in a zero-shot setting.

\begin{table}[t]
  \centering
  \setlength{\tabcolsep}{4pt}
  \begin{tabular}{lcccc}
    \toprule
    & CS $\uparrow$ & CT $\downarrow$ & RL $\uparrow$ & BSR $\uparrow$  \\
    \midrule
    Submission 1 & \underline{0.797} & \underline{0.696} & 0.283 & \underline{0.249} \\
    Submission 2 & 0.722 & 0.808 & 0.244 & 0.218 \\
    Submission 3 & 0.789 & 0.714 & \underline{0.292} & \textbf{0.295} \\
    Submission 4$^{*}$ & \textbf{0.818} & \textbf{0.621} & \textbf{0.305} & -  \\
    \midrule
    Official baseline 1 & 0.763 & 0.753 & 0.255 & 0.226 \\
    Official baseline 2 & 0.767 & 0.745 & 0.269 & 0.235 \\
    \bottomrule
  \end{tabular}
  \caption{Test results on Task 3.1. CS refers to consistency, CT refers to contradiction, RL refers to ROUGE recall, and BSR refers to BERTScore recall. The first baseline follows a zero-shot strategy using the model \texttt{LlamA-3.1-8B}. The second baseline adopts a pipeline approach where the predictions from the top performing baselines from Task 1 and 2 are used to enrich the prompt context. $^*$Submission 4 was made during the analysis phase of the competition.} 
  \label{tab:task3.1-results-test}
  \vspace{-0.3cm}
\end{table}

\subsection{Task 3.2}
We use \texttt{gpt} as the backbone LLM. The generated signatures are shown in Section~\ref{fig:signatures}. Our submission achieved a fit score of 0, a recurrence score of 0, and a specificity score of 0.25. Since this task was evaluated manually by the organizers and gold annotations are not available, it is challenging to provide findings for this task.

\section{Conclusion}
We presented our submissions to the CLPsych~2026 Shared Task on analyzing self-state dynamics in longitudinal Reddit timelines. Our ensembling approach for Task 1.1 achieved the best performance in subelement classification among all participants. Our pipeline approach for Task 2, which used predictions from Task 1, yielded our strongest result on this task and ranked fourth. The label-augmented in-context learning approach for Task 3.1 ranked third in the competition. Overall, our results indicate that propagating predictions from upstream tasks leads to consistent gains on downstream tasks.

\section*{Limitations}
Our work has several limitations. First, in our experiments, we used large LLMs that might not be accessible to everyone. This hinders the reproducibility and usage of the methods in resource-constrained environments. One solution is to use quantized models, which offer a reduced memory footprint at the cost of a slight degradation in performance. Second, the size of the datasets is relatively small and limited to 30 users in the training set and 10 users in the test set. Therefore, the conclusions from our paper might not be generalizable to the mental health domain in general. Third, the users in this study are Reddit users, who are not representative of the general population; to further extend the scope of the work, data from other social media platforms could be included. Finally, our submission to Task 3.2 received fit and recurrence scores of 0, suggesting that the batch-and-merge pipeline did not capture meaningful recurrent signatures; further investigation of the filtering and merging steps is needed.

\section*{Ethical Considerations}
    Due to the sensitive nature of mental health data, we stored the dataset in secure, firewall-protected servers. We used only open-weight LLMs and served them locally. We did not use any commercial, closed-weight LLMs in order to preserve the confidentiality of the data. Furthermore, our systems are designed as a support tool to help professional mental health providers and should not be considered a replacement for certified professionals.

\section*{Acknowledgment}
We thank Asmaa El Hansali for the help with the project and the anonymous reviewers for their insightful comments. 

\bibliography{custom}

\appendix

\section{Experimental Setup}
\label{sec:appendix-setup}

All experiments and inference are conducted on secure internal servers to protect the privacy of the individuals represented in the dataset. Our system is equipped with 8 NVIDIA A40 GPUs (48 GB VRAM each) and 40 CPU cores at 2.30 GHz. We serve all LLMs locally with vLLM \cite{kwon2023efficient}, a high-throughput inference engine for large language models.

Figure \ref{fig:example-timeline} presents an example timeline in the training dataset.

\section{Additional Results}
\label{sec:appendix-additional-results}
This section provides additional experiments and results.

\subsection{Task 1.1 and 1.2 submission details}
\label{appendix:sec-ensemble}

Tables~\ref{tab:ensemble-sub2} and~\ref{tab:ensemble-sub3} list the ensemble members for Submissions 2 and 3 in Task 1.1 and 1.2, respectively. All members use $k=3$.

\begin{table}[h]
\centering
\begin{tabular}{cll}
\toprule
\# & Approach & LLM \\
\midrule
1 & Post-level ICL with RAG & \texttt{gemma} \\
2 & Post-level ICL          & \texttt{gemma} \\
3 & Subelement-level ICL    & \texttt{gemma} \\
4 & Post-level ICL          & \texttt{qwen} \\
5 & Subelement-level ICL    & \texttt{qwen} \\
\bottomrule
\end{tabular}
\caption{Ensemble members for Submission 2 in Task 1.1 and 1.2. All members use $k=3$.}
\label{tab:ensemble-sub2}
\end{table}

\begin{table}[h]
\centering
\begin{tabular}{cll}
\toprule
\# & Approach & LLM \\
\midrule
1 & Post-level ICL with RAG & \texttt{gemma} \\
2 & Post-level ICL          & \texttt{gemma} \\
3 & Subelement-level ICL    & \texttt{gemma} \\
4 & Post-level ICL          & \texttt{qwen} \\
5 & Subelement-level ICL    & \texttt{qwen} \\
6 & Post-level ICL          & \texttt{gpt} \\
7 & Subelement-level ICL    & \texttt{gpt} \\
\bottomrule
\end{tabular}
\caption{Ensemble members for Submission 3 in Task 1.1 and 1.2. All members use $k=3$.}
\label{tab:ensemble-sub3}
\end{table}

\subsection{Task 2 experiments on the validation set}
\label{sec:hyperparams}
\paragraph{Training} While using the same training and validation set from Task~1, our model predicts F$_1$ scores generated by \texttt{gemma} and \texttt{qwen} across four prompting strategies---zero-shot (3 results from the \texttt{gemma} model and 2 results from \texttt{qwen}), post-level ICL (6 from each model), subelement-level ICL (6 from each model), and post-level RAG (6 from each model)--- yielding 41 training results. We evaluated four feature sets of increasing complexity (see Figures~\ref{fig:as_featureset}, \ref{fig:bs_featureset}, and~\ref{fig:esc_featureset}): FS1 comprises adaptive and maladaptive presence scores with their absolute inter-post deltas; FS2 adds the post index; FS3 adds per-post element counts; and FS4 combines FS3 with the post index. For both switch conditions (with foresight and without foresight), F$_1$ improved monotonically with feature set complexity, with FS4 yielding averages of $0.463$ and $0.472$ respectively. For escalation, FS3 produced a substantially stronger average F$_1$ of $0.706$ against $0.538$ and $0.541$ for FS1 and FS2, and was therefore selected.
 
We also evaluated window sizes $w_0$--$w_3$ (see Figures~\ref{fig:as_window}, \ref{fig:bs_window}, and~\ref{fig:esc_window}). Performance increased consistently with window size across all tasks, with $w_3$ achieving mean F$_1$ scores of $0.422$, $0.443$, and $0.607$ for switch with foresight, switch without foresight, and escalation, respectively; $w_3$ was therefore adopted. For the switch task, removing foresight---restricting the window to prior posts only---produced stronger predictions (F$_1$ $0.490$ vs.\ $0.472$ at FS4, $w_3$), and this formulation was retained for the final system. The best feature set configurations, FS4 for switch and FS3 for escalation, yield F$_1$ scores of $0.490$ and $0.714$, respectively.

\begin{table}[ht]
\centering
\small
\setlength{\tabcolsep}{4pt}
\renewcommand{\arraystretch}{1.8}
\begin{tabular}{@{}llll@{}}
\hline
\textbf{Sub.} & \textbf{Task} & \textbf{Model} & \textbf{Hyperparam.} \\
\hline
\multirow{2}{*}{2}
  & Escalation & \multirow{2}{*}{SVM} & \multirow{2}{3.9cm}{\texttt{C=1, kernel=`rbf', gamma=`scale'}} \\
  & Switch     &                      & \\
\hline
3 & Escalation & RF  & \multirow{2}{3.9cm}{\texttt{n\_est=200, depth=5, feat=`log2', split=5}} \\  \\
3 & Switch     & SVM & \multirow{2}{3.9cm}{\texttt{C=1, kernel=`rbf', gamma=`scale'}} \\ \\
\hline
\end{tabular}
\caption{Task~2 hyperparameters by submission and change type. RF = RandomForestClassifier; n\_est = n\_estimators, feat = max\_features; split = min\_samples\_split; depth = max\_depth}
\label{tab:task2-hyperparams}
\end{table}

\subsection{Task 3.1: zero-shot vs. summary of summaries}

We first tested a simple zero-shot prompt (Figure~\ref{fig:zeroshot}) 
in which the model receives the raw post texts and a brief instruction 
to generate a structured ABCD summary. This served as our baseline 
for both models.

For this section, we evaluate our approach on the Task 3.1 training set (74 sequences) 
using two language models: LLaMA 3.2 3B Instruct and Gemma 2 9B Instruct, 
both loaded with 4-bit quantization (NF4).

\paragraph{Sequential pipeline.}
Motivated by the winning system of \citet{sandu-etal-2025-capturing} (BLUE team), 
we implement a two-step sequential pipeline. 
In the first step, the model generates a short post-level summary 
for each individual post describing the interplay between adaptive 
and maladaptive self-states. In the second step, these post-level 
summaries -- rather than the raw post texts -- are fed to the model 
to produce the final sequence summary. The prompt used in the 
second step is shown in Figure~\ref{fig:sequential_prompt}.

\paragraph{Results.}
Table~\ref{tab:results} show the 
evaluation results across all configurations. The sequential pipeline 
consistently improved consistency over the zero-shot baseline for both models, 
with Gemma 2 9B achieving the best mean CS of 
$0.7382 \pm 0.0068$ across three runs.
 
\begin{table}[ht]
\centering
\begin{tabular}{llccccc}
\hline
\textbf{Sub.} & \textbf{Label} & \textbf{Prec} & \textbf{Rec} & \textbf{F$_1$} \\
\hline
\multirow{3}{*}{2}
  & Switch      & 0.500 & 0.381 & 0.432 \\
  & Escalation  & 0.591 & 0.542 & 0.565 \\
  & \textit{Macro F$_1$} & & & \textit{0.499} & & \\
\hline
\multirow{3}{*}{3}
  & Switch      & 0.326 & 0.714 & 0.448 \\
  & Escalation  & 0.625 & 0.833 & 0.714 \\
  & \textit{Macro F$_1$} & & & \textit{\textbf{0.581}} & & \\
\hline
\end{tabular}
\caption{Task 2: Post-level precision, recall, and F$_1$ by submission.}
\label{tab:post-level}
\end{table}
 
\begin{table}[ht]
\centering
\begin{tabular}{llccc}
\hline
\textbf{Sub.} & \textbf{Label} & \textbf{Prec} & \textbf{Rec} & \textbf{F$_1$} \\
\hline
\multirow{3}{*}{2}
  & Switch      & 0.450 & 0.342 & 0.349 \\
  & Escalation  & 0.675 & 0.508 & 0.542 \\
  & \textit{Macro F$_1$} & & & \textit{0.446} \\
\hline
\multirow{3}{*}{3}
  & Switch      & 0.353 & 0.592 & 0.416 \\
  & Escalation  & 0.714 & 0.721 & 0.709 \\
  & \textit{Macro F$_1$} & & & \textit{\textbf{0.563}} \\
\hline
\end{tabular}
\caption{Task 2: Timeline-level precision, recall, and F$_1$ (macro-averaged over 10 timelines) by submission.}
\label{tab:timeline-level}
\end{table}

\subsection{Task 3.1 results on validation set}

In addition to the approaches we discussed in the main text, we implement the following approaches:

\textbf{LLM as a judge.} Three independent ICL summaries are generated by \texttt{qwen}, \texttt{gemma}, and \texttt{gpt}. A separate LLM (\texttt{gpt}) then selects the best of the three.

\textbf{LLM as an aggregator.} Same setup as the judge, except that the aggregator LLM is asked to produce a new summary that draws on the three candidates rather than selecting one of them.

The results of the all approaches are reported in Table \ref{tab:task3.1-results-validation}.

\begin{table*}[th]
\centering
\setlength{\tabcolsep}{5pt}
\renewcommand{\arraystretch}{1.05}
\begin{tabular}{llcccc}
\toprule
Model & $k$ & CS $\uparrow$ & CT $\downarrow$ & RL $\uparrow$ & BSR $\uparrow$  \\
\midrule
\multirow{3}{*}{\texttt{qwen}}
 & 1 & 75.40 & 76.96 & 25.47 & 28.63 \\
 & 2 & 75.95 & 75.93 & 26.15 & 30.11 \\
 & 3 & 76.20 & 75.33 & 27.45 & 31.49 \\
\midrule
\multirow{3}{*}{\texttt{gpt}}
 & 1 & 77.88 & 70.60 & 27.90 & 27.43 \\
 & 2 & 79.68 & \textbf{66.14} & 28.56 & 26.59 \\
 & 3 & \textbf{80.56} & \underline{67.71} & \underline{28.59} & 27.85 \\
\midrule
\multirow{3}{*}{\texttt{gemma}}
 & 1 & 75.23 & 75.32 & 26.40 & 31.59 \\
 & 2 & 75.83 & 73.09 & 28.51 & \underline{33.41} \\
 & 3 & 74.78 & 75.16 & \textbf{28.84} & \textbf{34.10} \\
\midrule
\multirow{3}{*}{\texttt{aggregate}}
 & 1 & \underline{79.96} & 70.26 & 27.82 & 26.59 \\
 & 2 & 78.61 & 69.89 & 28.56 & 26.57 \\
 & 3 & 78.38 & 71.01 & 28.36 & 27.39 \\
\midrule
\multirow{3}{*}{\texttt{judge}}
 & 1 & 76.01 & 72.50 & 27.19 & 29.34 \\
 & 2 & 76.58 & 73.10 & 27.23 & 28.37 \\
 & 3 & 77.74 & 71.86 & 27.47 & 27.22 \\
\midrule
\multirow{3}{*}{\texttt{simple-prompt}}
 & 1 & 77.99 & 69.75 & 28.09 & 27.37 \\
 & 2 & 77.73 & 71.70 & 28.03 & 26.60 \\
 & 3 & 79.43 & 68.45 & 28.50 & 27.77 \\
\bottomrule
\end{tabular}
\caption{Average CS (consistency), CT (contradiction), RL (ROUGE recall), and BSR (BERTScore recall) over 4 independent runs on the validation data. All the numbers are in percentages. The first three models represent ICL methods with different LLMs. In \texttt{aggregate} (LLM as an aggregator) and \texttt{judge} (LLM as a judge), \texttt{gpt} uses the predicitons form the first three models. The \texttt{short-prompt} model is ICL with \texttt{gpt} as the backbone but with a shorter prompt (Figure \ref{fig:sys-prompt-3.1-short}). Ensemble methods for summary generation degrade the performance of ICL methods.}
\label{tab:task3.1-results-validation}
\end{table*}

\section{Task 3.2 Recurrent Signatures}
\label{fig:signatures}

Here is the detected recurrent signatures of change by our method:

\paragraph{Signature of Deterioration.}
The recurrent signature of deterioration is characterized by a shift 
from adaptive self-states marked by (C-S) and (D) to maladaptive 
self-states dominated by (A), (C-S), (C-O), and (D). Initially, 
adaptive self-states are present, often characterized by (A) and (D), 
but these are gradually overshadowed by maladaptive states. The 
maladaptive states intensify (A), (C-S), (C-O), and (D) mutually 
reinforcing each other, culminating in a sense of hopelessness and despair.

\paragraph{Signature of Improvement.}
The recurrent signature of improvement is characterized by a shift 
from maladaptive self-criticism (C-S) and depressive affect (A) to 
adaptive self-compassion (C-S) and content affect (A). This shift 
involves maladaptive self-neglect behaviors (B-S) being overshadowed 
by relating behaviors (B-O) and a strengthened desire for connection 
(D) mutually reinforcing adaptive self-compassion (C-S).

\begin{table*}[h]
\centering
\begin{tabular}{llccc}
\hline
\textbf{Configuration} & \textbf{Model} & \textbf{CS} $\uparrow$ & \textbf{CT} $\downarrow$ & \textbf{ROUGE-L} $\uparrow$ \\
\hline
\multirow{2}{*}{Zero-shot baseline}
  & LLaMA 3.2 3B & 0.7152 & 0.6233 & 0.1502 \\
  & Gemma 2 9B   & 0.7086 & 0.7920 & 0.1837 \\
\hline
\multirow{4}{*}{Sequential pipeline}
  & LLaMA 3.2 3B -- Run 1 & 0.7275 & 0.7187 & 0.2135 \\
  & LLaMA 3.2 3B -- Run 2 & 0.7279 & 0.6979 & 0.2117 \\
  & LLaMA 3.2 3B -- Run 3 & 0.7202 & 0.7106 & 0.2076 \\
  & \textbf{LLaMA mean $\pm$ std} & $\mathbf{0.7252 \pm 0.0040}$ & -- & -- \\
\hline
\multirow{4}{*}{Sequential pipeline}
  & Gemma 2 9B -- Run 1 & 0.7408 & 0.7644 & 0.2006 \\
  & Gemma 2 9B -- Run 2 & 0.7304 & 0.7463 & 0.1900 \\
  & Gemma 2 9B -- Run 3 & 0.7433 & 0.7674 & 0.1906 \\
  & \textbf{Gemma mean $\pm$ std} & $\mathbf{0.7382 \pm 0.0068}$ & -- & -- \\
\hline
\end{tabular}
\caption{Task 3.1 Evaluation Results on Training Set (74 sequences).}
\label{tab:results}
\end{table*}

%
%
\begin{figure*}[t]
  \centering

  \begin{subfigure}[t]{0.48\linewidth}
    \centering
    \begin{tikzpicture}
      \begin{axis}[
        barfig_std,
        scale=0.65,
        xtick=data,
        symbolic x coords={w0, w1, w2, w3},
        xlabel={Window Size},
      ]
        \addplot[fill={rgb,255:red,76;green,114;blue,176},
                 draw={rgb,255:red,50;green,80;blue,130}]
          coordinates {(w0,0.4055) (w1,0.4380) (w2,0.4128) (w3,0.4222)};
      \end{axis}
    \end{tikzpicture}
    \caption{Window size effect.}
    \label{fig:as_window}
  \end{subfigure}
  \hfill
  \begin{subfigure}[t]{0.48\linewidth}
    \centering
    \begin{tikzpicture}
      \begin{axis}[
        barfig_std,
        scale=0.65,
        xtick=data,
        symbolic x coords={FS1, FS2, FS3, FS4},
        xlabel={Feature Set},
      ]
        \addplot[fill={rgb,255:red,221;green,132;blue,82},
                 draw={rgb,255:red,180;green,100;blue,50}]
          coordinates {(FS1,0.3927) (FS2,0.3792) (FS3,0.4441) (FS4,0.4627)};
      \end{axis}
    \end{tikzpicture}
    \caption{Feature set effect.}
    \label{fig:as_featureset}
  \end{subfigure}

  \caption*{\textbf{Switch with Foresight (AS)}}

  \medskip

  \begin{subfigure}[t]{0.48\linewidth}
    \centering
    \begin{tikzpicture}
      \begin{axis}[
        barfig_std,
        scale=0.65,
        xtick=data,
        symbolic x coords={w0, w1, w2, w3},
        xlabel={Window Size},
      ]
        \addplot[fill={rgb,255:red,76;green,114;blue,176},
                 draw={rgb,255:red,50;green,80;blue,130}]
          coordinates {(w0,0.4201) (w1,0.4283) (w2,0.4346) (w3,0.4425)};
      \end{axis}
    \end{tikzpicture}
    \caption{Window size effect.}
    \label{fig:bs_window}
  \end{subfigure}
  \hfill
  \begin{subfigure}[t]{0.48\linewidth}
    \centering
    \begin{tikzpicture}
      \begin{axis}[
        barfig_std,
        scale=0.65,
        xtick=data,
        symbolic x coords={FS1, FS2, FS3, FS4},
        xlabel={Feature Set},
      ]
        \addplot[fill={rgb,255:red,221;green,132;blue,82},
                 draw={rgb,255:red,180;green,100;blue,50}]
          coordinates {(FS1,0.3979) (FS2,0.4145) (FS3,0.4410) (FS4,0.4722)};
      \end{axis}
    \end{tikzpicture}
    \caption{Feature set effect.}
    \label{fig:bs_featureset}
  \end{subfigure}

  \caption*{\textbf{Switch without Foresight (BS)}}

  \medskip

  \begin{subfigure}[t]{0.48\linewidth}
    \centering
    \begin{tikzpicture}
      \begin{axis}[
        barfig_esc,
        scale=0.65,
        xtick=data,
        symbolic x coords={w1, w2, w3},
        xlabel={Window Size},
      ]
        \addplot[fill={rgb,255:red,76;green,114;blue,176},
                 draw={rgb,255:red,50;green,80;blue,130}]
          coordinates {(w1,0.5851) (w2,0.5927) (w3,0.6068)};
      \end{axis}
    \end{tikzpicture}
    \caption{Window size effect (w0 not evaluated).}
    \label{fig:esc_window}
  \end{subfigure}
  \hfill
  \begin{subfigure}[t]{0.48\linewidth}
    \centering
    \begin{tikzpicture}
      \begin{axis}[
        barfig_esc,
        scale=0.65,
        xtick=data,
        symbolic x coords={FS1, FS2, FS3},
        xlabel={Feature Set},
      ]
        \addplot[fill={rgb,255:red,221;green,132;blue,82},
                 draw={rgb,255:red,180;green,100;blue,50}]
          coordinates {(FS1,0.5377) (FS2,0.5412) (FS3,0.7056)};
      \end{axis}
    \end{tikzpicture}
    \caption{Feature set effect (FS4 not evaluated).}
    \label{fig:esc_featureset}
  \end{subfigure}

  \caption*{\textbf{Escalation (ESC)}}

  \caption{Mean F1 scores per window size (left column) and feature set (right column)
    for each task. FS1: presence + absolute deltas; FS2: FS1 + Post Index;
    FS3: FS1 + count features; FS4: FS3 + Post Index.}
  \label{fig:all_results}
\end{figure*}

\begin{table*}[t]
  \centering
  \setlength{\tabcolsep}{4pt}
  \begin{tabular}{l ccc ccc c}
    \toprule
    & \multicolumn{3}{c}{\textbf{Post-Level}} & \multicolumn{3}{c}{\textbf{Timeline-Level}} & \\
    \cmidrule(lr){2-4} \cmidrule(lr){5-7}
    & \multicolumn{1}{c}{Switch F1} & \multicolumn{1}{c}{Esc.\ F1} & \multicolumn{1}{c}{Macro F1} 
    & \multicolumn{1}{c}{Switch F1} & \multicolumn{1}{c}{Esc.\ F1} & \multicolumn{1}{c}{Macro F1} 
    & \multicolumn{1}{c}{Combined F1 $\uparrow$} \\
    \midrule
    Submission 1 & 0.304 & 0.588 & 0.446 & 0.239 & 0.480 & 0.359 & 0.403 \\
    Submission 2 & 0.346 & 0.507 & 0.426 & 0.344 & 0.319 & 0.331 & 0.379 \\
    Submission 3 & \textbf{0.510} & \textbf{0.791} & \textbf{0.650} & \textbf{0.559} & \textbf{0.697} & \textbf{0.628} & \textbf{0.639} \\
    \bottomrule
  \end{tabular}
  \caption{Validation results on Task 2. Post-level and timeline-level scores are reported per label and as macro F1; the combined F1 is the ranking metric.}
  \label{tab:2-val-results}
  \vspace{-0.3cm}
\end{table*}

\section{Dataset}
\label{sec:appendix-dataset}

Figure \ref{fig:example-timeline} presents an example timeline in the training dataset.

\section{Prompts}
\label{sec:appendix-prompts}
Figures \ref{fig:sys-prompt-1.1}, \ref{fig:zeroshot}, \ref{fig:sequential_prompt}, \ref{fig:sys-prompt-3.1}, and \ref{fig:sys-prompt-3.1-short} show the system prompts we used in our approaches.

\begin{figure*}[!t]
\begin{lstlisting}
{
  "timeline_id": "7d9d2e0e0a",
  "posts": [
    {
      "post_index": 1,
      "post_id": "a33649e870",
      "date": "01-01-2020, 01:02:03",
      "Switch": "0",
      "Escalation": "E",
      "post": "I'm tired. [REMOVED]",
      "Well-being": 4,
      "evidence": {
        "adaptive-state": {
          "B-O": {
            "Category": "(1) Relating behavior",
            "highlighted_evidence": "[REMOVED]"
          },
          "Presence": 2
        },
        "maladaptive-state": {
          "A": {
            "Category": "(4) Depressed, despair, hopeless",
            "highlighted_evidence": "[REMOVED]"
          },
          "B-S": {
            "Category": "(2) Self harm, neglect and avoidance",
            "highlighted_evidence": "[REMOVED]"
          },
          "Presence": 5
        }
      }
    },
    {
      "post_index": 2,
      "post_id": "7f22000b69",
      "date": "01-01-2021, 01:02:03",
      "Switch": "S",
      "Escalation": "E",
      "post": "I'm tired of being alone [REMOVED]",
      "Well-being": 1,
      "evidence": {
        "adaptive-state": {
          "C-O": {
            "Category": "(1) Perception of the other as related",
            "highlighted_evidence": "[REMOVED]"
          },
          "Presence": 2
        },
        "maladaptive-state": {
          "A": {
            "Category": "(4) Depressed, despair, hopeless",
            "highlighted_evidence": "[REMOVED]"
          },
          "Presence": 5
        }
      }
    }
  ]
}
\end{lstlisting}
    \caption{A sample timeline data.}
    \label{fig:example-timeline}
\end{figure*}

\begin{figure*}[t]
\begin{promptbox}
## Task:
Your task is to identify adaptive and maladaptive self-states from a reddit post. A post can contain zero or more adaptive and maladaptive self-states.

### Definitions
Self-states are conceptualized as structured combinations of 6 elements: Affect (A), Behavior of the self with the others (B-O), Behavior of the self toward the self (B-S), Cognition of the others (C-O), Cognition of the self (C-S), and Desire (D). Each present element has exactly one subelement. The subelements are the specific manifestations of the elements in the post. Here are the definitions of the elements and their subelements. The percentage of subelements across all posts in the training data is provided in parentheses:
- *Affect* (A): Emotional tone or mood.
    - Adaptive subelements:
        1. Calm / laid back (0.60%)
        2. Sad, emotional pain, grieving (5.36%)
        3. Content, happy, joy, hopeful (8.33%)
        4. Vigor / energetic (0.60%)
        5. Justifiable anger/ assertive anger, justifiable outrage (1.19%)
        6. Proud (4.17%)
        7. Feeling loved, belong (0.00%)
    - Maladaptive subelements:
        1. Anxious/ fearful/ tense (16.67%)
        2. Depressed, despair, hopeless (34.52%)
        3. Mania (0.60%)
        4. Apathic, don't care, blunted (0.00%)
        5. Angry (aggression), disgust, contempt (4.76%)
        6. Ashamed, guilty (4.76%)
        7. Feel lonely (4.76%)
- *Behavior of the self with the others* (B-O): The writer's main behavior(s) toward the others.
    - Adaptive subelements:
        1. Relating behavior (48.21%)
        2. Autonomous or adaptive control behavior
    - Maladaptive subelements:
        1. Fight or flight behavior (13.69%)
        2. Over-controlled or controlling behavior
- *Behavior toward the self* (B-S): The writer's main behavior(s) toward the self.
    - Adaptive subelements:
        1. Self care and improvement (34.52%)
    - Maladaptive subelements:
        1. Self harm, neglect and avoidance (27.98%)
- *Cognition of the others* (C-O): The writer's main perceptions of the other.
    - Adaptive subelements:
        1. Perception of the other as related (19.05%)
        2. Perception of the other as facilitating autonomy needs (1.19%)
    - Maladaptive subelements:
        1. Perception of the other as detached or over attached (45.83%)
        2. Perception of the other as blocking autonomy needs (6.55%)
- *Cognition of the self* (C-S): The writer's main self-perceptions.
    - Adaptive subelements:
        1. Self-acceptance and compassion (25.00%)
    - Maladaptive subelements:
        1. Self criticism (57.14%)
- *Desire* (D): The writer's main desire, expectation, need, intention, or fear.
    - Adaptive subelements:
        1. Relatedness (24.40%)
        2. Autonomy and adaptive control (7.14%)
        3. Competence, self esteem, self-care (21.43%)
    - Maladaptive subelements:
        1. Expectation that relatedness needs will not be met (13.10%)
        2. Expectation that autonomy needs will not be met (4.76%)
        3. Expectation that competence needs will not be met (26.19%)

Self-state rating is the degree to which each identified self state is present in the post. It is an integer between 1 and 5 with the following definitions:
- 1 (Not present): The self state is not expressed in the post.
- 2 (Somewhat present): The self state is expressed, but plays a subtle, limited role in shaping the person's overall experience.
- 3 (Moderately present): The self state is clearly expressed and moderately contributes to the person's experience.
- 4 (Much present): The self state strongly influences and shapes the experience described in the post.
- 5 (Highly present): The self state strongly shapes and clearly defines the overall experience described in the post.
...
\end{promptbox}
    \caption{The system prompt for the Task 1.1 and 1.2.}
    \label{fig:sys-prompt-1.1}
\end{figure*}

\begin{figure*}
\begin{promptbox}
...
Percentage of subelements across all posts in the training data is provided in parentheses:
- Affect (A):
    - Adaptive subelements: 1 (0.60%), 2 (5.36%), 3 (8.33%), 4 (0.60%), 5 (1.19%), 6 (4.17%), 7 (0.00%)
    - Maladaptive subelements: 1 (16.67%), 2 (34.52%), 3 (0.60%), 4 (0.00%), 5 (4.76%), 6 (4.76%), 7 (4.76%)
- Behavior of the self with the others (B-O):
    - Adaptive subelements: 1 (48.21%), 2 (4.76%)
    - Maladaptive subelements: 1 (13.69%), 2 (0.00%)
- Behavior toward the self (B-S):
    - Adaptive subelements: 1 (34.52%)
    - Maladaptive subelements: 1 (27.98%)
- Cognition of the others (C-O):
    - Adaptive subelements: 1 (19.05%), 2 (1.19%)
    - Maladaptive subelements: 1 (45.83%), 2 (6.55%)
- Cognition of the self (C-S):
    - Adaptive subelements: 1 (25.00%)
    - Maladaptive subelements: 1 (57.14%)
- Desire (D):
    - Adaptive subelements: 1 (24.40%), 2 (7.14%), 3 (21.43%)
    - Maladaptive subelements: 1 (13.10%), 2 (4.76%), 3 (26.19%)


### Output format
You need to output the presence and rating of the adaptive and maladaptive self-states as well as the subelements of the self-states. The subelement of non-existent self-states should be 0. Write your output in the following JSON format:
```json
{
    "adaptive_states": {
        "A": int (integer between 0 and 7),
        "B-O": int (integer between 0 and 2),
        "B-S": int (integer between 0 and 1),
        "C-O": int (integer between 0 and 2),
        "C-S": int (integer between 0 and 1),
        "D": int (integer between 0 and 3),
        "rating": int (integer between 1 and 5)
    },
    "maladaptive_states": {
        "A": int (integer between 0 and 7),
        "B-O": int (integer between 0 and 2),
        "B-S": int (integer between 0 and 1),
        "C-O": int (integer between 0 and 2),
        "C-S": int (integer between 0 and 1),
        "D": int (integer between 0 and 3),
        "rating": int (integer between 1 and 5)
    }
}
```
\end{promptbox}
    \caption{The system prompt for the Task 1.1 and 1.2. (Cont.)}
\end{figure*}

\begin{figure*}
\begin{promptbox}
...
SYSTEM_PROMPT = (
    'You are a clinical psychologist specialising in psychodynamic self-state analysis.\n'
    'Your task is to write a structured sequence summary for social media posts '
    'surrounding a mental health change event, grounded in the MIND (ABCD) framework.\n\n'
    'FRAMEWORK:\n'
    'Self-states combine: Affect (A), Behavior-Self (B-S), Behavior-Other (B-O), '
    'Cognition-Self (C-S), Cognition-Other (C-O), Desire (D).\n'
    'Each self-state is Adaptive or Maladaptive. Always abbreviate ABCD in parentheses.\n\n'
    'OUTPUT REQUIREMENTS (in this order, up to 350 words):\n'
    '1. CENTRAL THEME: Dominant ABCD pattern; direction of change '
    '(improvement/deterioration); change event type (Switch/Escalation/both); when it occurs.\n'
    '2. ADAPTIVE DYNAMICS: Presence trajectory and internal ABCD relational dynamics.\n'
    '3. MALADAPTIVE DYNAMICS: Same for the maladaptive state.\n'
    '4. CROSS-STATE DYNAMICS: Dominance, suppression, or dialogue between states.\n\n'
    'CONSTRAINTS: Do NOT print numeric presence scores. Max 350 words.'
)
...
\end{promptbox}
    \caption{Simple zero-shot system prompt for Task3.1 used as baseline for both LLaMA 3.2 3B and Gemma 2 9B.}
    \label{fig:zeroshot}
\end{figure*}

\begin{figure*}
\begin{promptbox}
...
{Step 1 -- Post-level prompt (applied to each post individually):}

{``Summarise the interplay between adaptive and maladaptive 
self-states in this single post. Identify the dominant self-state 
and describe how the core ABCD elements interact. 
Write 2--3 sentences only.''}

{Step 2 -- Sequence-level prompt (applied to all post summaries):}

You are a clinical psychologist specialising in psychodynamic 
self-state analysis. Your task is to write a structured sequence 
summary grounded in the MIND (ABCD) framework.

{Change event definitions:}
SWITCH: sudden change in well-being between two consecutive posts.
ESCALATION: gradual intensification of mood across consecutive posts.

'OUTPUT REQUIREMENTS (up to 350 words, one paragraph):\n'
    '1. CENTRAL THEME: dominant ABCD pattern; direction 
    (improvement/deterioration); change event type (Switch/Escalation)
    '2. ADAPTIVE DYNAMICS: presence trajectory; relational dynamics 
    between ABCD subelements (MUST be described)
    '3. MALADAPTIVE DYNAMICS: same for the maladaptive state
    '4. CROSS-STATE DYNAMICS: dominance, suppression, reflective 
    dialogue (MUST be described if present)
    'CONSTRAINTS: Do NOT print numeric presence scores. Max 350 words.'
    
{Constraints:} Use (A),(B-S),(B-O),(C-S),(C-O),(D) abbreviations. 
No numeric scores. Start with: 
{``The central psychological theme revolves around''}
...
\end{promptbox}
    \caption{Sequential pipeline prompt for Task3.1 following \citet{sandu-etal-2025-capturing}. 
Step 1 generates post-level summaries; Step 2 generates the sequence 
summary from those summaries rather than raw post texts.}
    \label{fig:sequential_prompt}
\end{figure*}

\begin{figure*}
\begin{promptbox}
You are a clinical psychologist specialising in psychodynamic self-state analysis.
Your task is to write a structured sequence summary grounded in the MIND (ABCD) framework.

# Framework
In the MIND framework, a self-state is defined as an identifiable unit characterized by specific combinations of Affect, Behavior (towards the self and others), Cognition (towards the self and others),and Desire (ABCD). An Adaptive self-state pertains to aspects of ABCD that are conducive to the fulfillment of basic desires/needs. A Maladaptive self-state pertains to aspects of ABCD that hinder the fulfillment of basic desires/needs. Each of the ABCD elements is operationalized through a set of subelements, representing distinct psychological expressions within each element:

1. *Affect* (A): Emotional tone or mood
    - Adaptive subelements: Calm/ laid back; Sad, emotional pain, grieving; Content, happy, joy, hopeful; Vigor / energetic; Justifiable anger/ assertive anger, justifiable outrage; Proud; Feeling loved, belong
    - Maladaptive subelements: Anxious/ fearful/ tense; Depressed, despair, hopeless; Mania; Apathic, don't care, blunted; Angry (aggression), disgust, contempt; Ashamed, guilt; Feel lonely
2. *Behavior of the self with the others* (B-O): The writer's main behavior(s) toward the others
    - Adaptive subelements: Relating behavior; Autonomous or adaptive control behavior
    - Maladaptive subelements: Fight or flight behavior; Over-controlled or controlling behavior
3. *Behavior toward the self* (B-S): The writer's main behavior(s) toward the self
    - Adaptive subelements: Self care and improvement
    - Maladaptive subelements: Self harm, neglect and avoidance
4. *Cognition of the others* (C-O): The writer's main perceptions of the other
    - Adaptive subelements: Perception of the other as related; Perception of the other as facilitating autonomy needs
    - Maladaptive subelements: Perception of the other as detached or over attached; Perception of the other as blocking autonomy needs
5. *Cognition of the self* (C-S): The writer's main self-perceptions
    - Adaptive subelements: Self-acceptance and compassion
    - Maladaptive subelements: Self criticism
6. *Desire* (D): The writer's main desire, expectation, need, intention, or fear.
    - Adaptive subelements: Relatedness; Autonomy and adaptive control; Competence, self esteem, self-care
    - Maladaptive subelements: Expectation that relatedness needs will not be met; Expectation that autonomy needs will not be met; Expectation that competence needs will not be met

Given a chronologically ordered sequence of posts from a single individual (a timeline), there are two possible clinically meaningful moments of change:

- SWITCH:  A switch reflects a substantial and sudden change in well-being between two consecutive posts. The change may reflect either improvement or deterioration.
- ESCALATION:  An escalation refers to a gradual intensification of mood over a sequence of consecutive posts. It occurs when an individual's mood progressively shifts from neutral or mildly valenced, toward a more extreme state. An escalation may reflect either improvement or deterioration.

# Task
Your task is to generate a structured summary describing patterns of self-state dynamics and their progression over time within a sequence of posts surrounding a change (Switch or Escalation). The summary must describe how psychological change processes evolve across the sequence, and how they culminate in (when it's a Switch), or unfold through (when it's an Escalation), the identified change event. The direction of the change (improvement / deterioration) as well as the identity of the change event (Switch/Escalation) should be explicitly stated in the summary. Describe the change pattern using the MIND framework (ABCD elements).

The summary should include references, only when they are evident in the data, to the following aspects:
1. *Central recurring theme across the posts*: Describe the central dynamic psychological theme and change trajectory characterizing the change process across the sequence. Explain how this theme evolves across the sequence. The theme should be described across the stages of the change process within the sequence, making clear how the theme appears before the change and how it develops as the change unfolds or when it culminates.
2. *Dynamics within the Adaptive and Maladaptive self-states and their presence*: Describe how present each self-state is and how its relative presence changes throughout the sequence as part of the change process. Presence refers to how strongly each self-state is expressed or dominant at different points in the sequence, whereas dynamics refer to the interactions between the ABCD subelements within that self-state. Where present, describe the adaptive and/or the maladaptive self-states in terms of ABCD subelements through explicit relational dynamics between them within the same self-state. If a self-state is described, relational dynamics between its ABCD subelements MUST also be described. Dynamics within a self-state are relational patterns between two or more subelements within the same self-state. These dynamics may be directional or reciprocal, such as co-activation, mutual reinforcement, exacerbation of one element by another, amplification of one element by another, or other structured interactions.
3. *Relationship between the adaptive and maladaptive self states and their relative presence*: Describe how the adaptive and maladaptive self-states relate to one another and how that changes throughout the sequence. Describe how the relative presence and dominance of the adaptive and maladaptive self-states shifts across the sequence. This may include: one self-state dominating the other, suppressing or silencing the other, or both self-states coexisting through reflective dialogue. examine whether dynamics occur between ABCD subelements across opposite self-states (suppression/attenuation, reflective dialogue, dominance competition, resilience or other structured interactions). If such cross-self-state dynamics are present in the sequence, they MUST be described.

# Output requirement
Each reference to an ABCD element should include its abbreviation in parentheses. Use the following mapping: (A) for Affect; (B-S) for Behavior-self; (B-O) for Behavior-other; (C-S) for Cognition-self; (C-O) for Cognition-other; (D) for Desire. Keep your summary below 350 words and write it ONE paragraph. Don not mention the post numbers in your summary.

Follow the structre of the examples and write your summary in the same format and start with the term "The central psychological theme revolves around..."

Think step by step and analyze the posts. Finally, write your summary in the following format:
```json
{
	"summary": "<YOUR SUMMARY>"
}
```
\end{promptbox}
    \caption{The system prompt for the Task 3.1.}
    \label{fig:sys-prompt-3.1}
\end{figure*}

\begin{figure*}[ht]
\begin{promptbox}
You are a clinical psychologist specializing in psychodynamic self-state analysis. Write a structured sequence summary grounded in the MIND (ABCD) framework.

# Framework
A self-state is characterized by combinations of Affect (A), Behavior toward others (B-O), Behavior toward self (B-S), Cognition of others (C-O), Cognition of self (C-S), and Desire (D). Self-states are either Adaptive (ABCD elements conducive to fulfilling basic needs) or Maladaptive (ABCD elements that hinder need fulfillment).

Subelements:
- A - Adaptive: calm, sad/grieving, content/hopeful, vigor, justifiable anger, proud, feeling loved. Maladaptive: anxious, depressed/hopeless, manic, apathetic, aggression/contempt, ashamed, lonely.
- B-O - Adaptive: relating, autonomous/adaptive control. Maladaptive: fight/flight, over-controlling.
- B-S - Adaptive: self-care/improvement. Maladaptive: self-harm/neglect/avoidance.
- C-O - Adaptive: other as related, other as autonomy-facilitating. Maladaptive: other as detached/over-attached, other as blocking autonomy.
- C-S - Adaptive: self-acceptance/compassion. Maladaptive: self-criticism.
- D - Adaptive: relatedness, autonomy, competence/self-esteem. Maladaptive: expectation that relatedness, autonomy, or competence needs won't be met.

Change events:
- SWITCH: Sudden, substantial change in well-being between two consecutive posts.
- ESCALATION: Gradual intensification of mood across consecutive posts toward a more extreme state.
Both can reflect improvement or deterioration.

# Task
Write a structured summary describing self-state dynamics and their progression across a chronological post sequence surrounding a change event. Cover:
1. Central recurring theme - how it evolves before and through/culminating in the change.
2. Adaptive and maladaptive self-state dynamics - their relative presence and ABCD subelement interactions (co-activation, reinforcement, amplification, etc.) across the sequence.
3. Relationship between self-states - how dominance, suppression, coexistence, or cross-state dynamics shift across the sequence.

Explicitly state the change direction (improvement/deterioration) and type (Switch/Escalation). Include ABCD abbreviations inline. Stay under 350 words, one paragraph, starting with: "The central psychological theme revolves around..."

Output format:
{
  "summary": "<YOUR SUMMARY>"
}
\end{promptbox}
    \caption{The shorter version of system prompt for the Task 3.1.}
    \label{fig:sys-prompt-3.1-short}
\end{figure*}

\end{document}